\documentclass[10pt,conference,a4paper]{IEEEtran}
\usepackage{cite}

\ifCLASSINFOpdf
  \usepackage[pdftex]{graphicx}
  \graphicspath{{Images/}}
  \DeclareGraphicsExtensions{.pdf,.jpeg,.png}
\else
\fi
\usepackage{amsmath}
\usepackage{array}

\ifCLASSOPTIONcompsoc
  \usepackage[caption=false,font=normalsize,labelfont=sf,textfont=sf]{subfig}
\else
  \usepackage[caption=false,font=footnotesize]{subfig}
\fi
\usepackage{hyperref}
\usepackage{url}

\usepackage{textcomp} 
\usepackage{gensymb} 

\usepackage{adjustbox}

\usepackage{fancyhdr}

\begin{document}
%
\title{Multimodal End-to-End Learning for Autonomous Steering in Adverse Road and Weather Conditions}


%
\author{\IEEEauthorblockN{Jyri Maanp{\"a}{\"a}\IEEEauthorrefmark{1}\IEEEauthorrefmark{2},
Josef Taher\IEEEauthorrefmark{1},
Petri Manninen\IEEEauthorrefmark{1}, 
Leo Pakola\IEEEauthorrefmark{1},
Iaroslav Melekhov\IEEEauthorrefmark{2} and
Juha Hyypp{\"a}\IEEEauthorrefmark{1}}
\IEEEauthorblockA{\IEEEauthorrefmark{1}Department of Remote Sensing and Photogrammetry, Finnish Geospatial Research Institute FGI, \\ National Land Survey of Finland, 02430 Masala, Finland, email: jyri.maanpaa@nls.fi}
\IEEEauthorblockA{\IEEEauthorrefmark{2}Aalto University School of Science, 02150 Espoo, Finland}}


\maketitle

\begin{abstract}
	Autonomous driving is challenging in adverse road and weather conditions in which there might not be lane lines, the road might be covered in snow and the visibility might be poor. We extend the previous work on end-to-end learning for autonomous steering to operate in these adverse real-life conditions with multimodal data. We collected 28 hours of driving data in several road and weather conditions and trained convolutional neural networks to predict the car steering wheel angle from front-facing color camera images and lidar range and reflectance data. We compared the CNN model performances based on the different modalities and our results show that the lidar modality improves the performances of different multimodal sensor-fusion models.
	We also performed on-road tests with different models and they support this observation.
\end{abstract}


%
\IEEEpeerreviewmaketitle

\section{Introduction}

\thispagestyle{fancy}
Autonomous driving is challenging in off-road tracks and adverse weather conditions, including fog, snowfall and heavy rain. These problems present a challenge for choosing the appropriate sensor suite and perception methods for autonomous cars to operate in these conditions. As these road and weather conditions are common in several areas, these problems need to be solved for self-driving cars to become fully autonomous. 

Current autonomous driving research has a focus on the application of different sensor types for autonomous cars and sensor fusion to operate the car based on this multimodal data. Two of the most applied sensor types for sensor fusion are RGB camera and lidar, which provide dense data with rich information about the car surroundings. However, it requires well generalized pattern recognition methods in order to detect all of the important objects and to filter all redundancy in the data.

Convolutional neural networks (CNNs) are widely used in many image analysis tasks and they are the state of the art in several image-based pattern recognition problems. 
Based on deep learning in their modern form, these methods require large training datasets of observed problem scenarios for the models to gain a general internal representation of the input data. CNNs can generalize their operation due to their convolutional structure enforcing general behaviour, but they need to be trained in a way that prevents overfitting effects. 

In adverse road and weather conditions, the detection of the road and the lane is an important task due to the high variance in the road environment.
In the work of M. Bojarski et al.~\cite{bojarski2016end}, a CNN was trained with car front camera data to predict the steering angle of the car in an end-to-end behaviour cloning fashion. The CNN could be operated online to steer the car in several road environments and weather conditions, including gravel road, rain and light snow based on their video footage\footnote{\url{https://www.youtube.com/watch?v=NJU9ULQUwng&t=315}}. Their model reportedly reached an approximately~98\% level of autonomy (fraction of autonomously driven time of test time), with a supposedly maximum distance of 10 miles with no intercepts in highway environment. The model operated only with the intention to stay on the current lane, excluding turns in intersections.

\begin{figure}[!t]
\centering
\includegraphics[width=\linewidth]{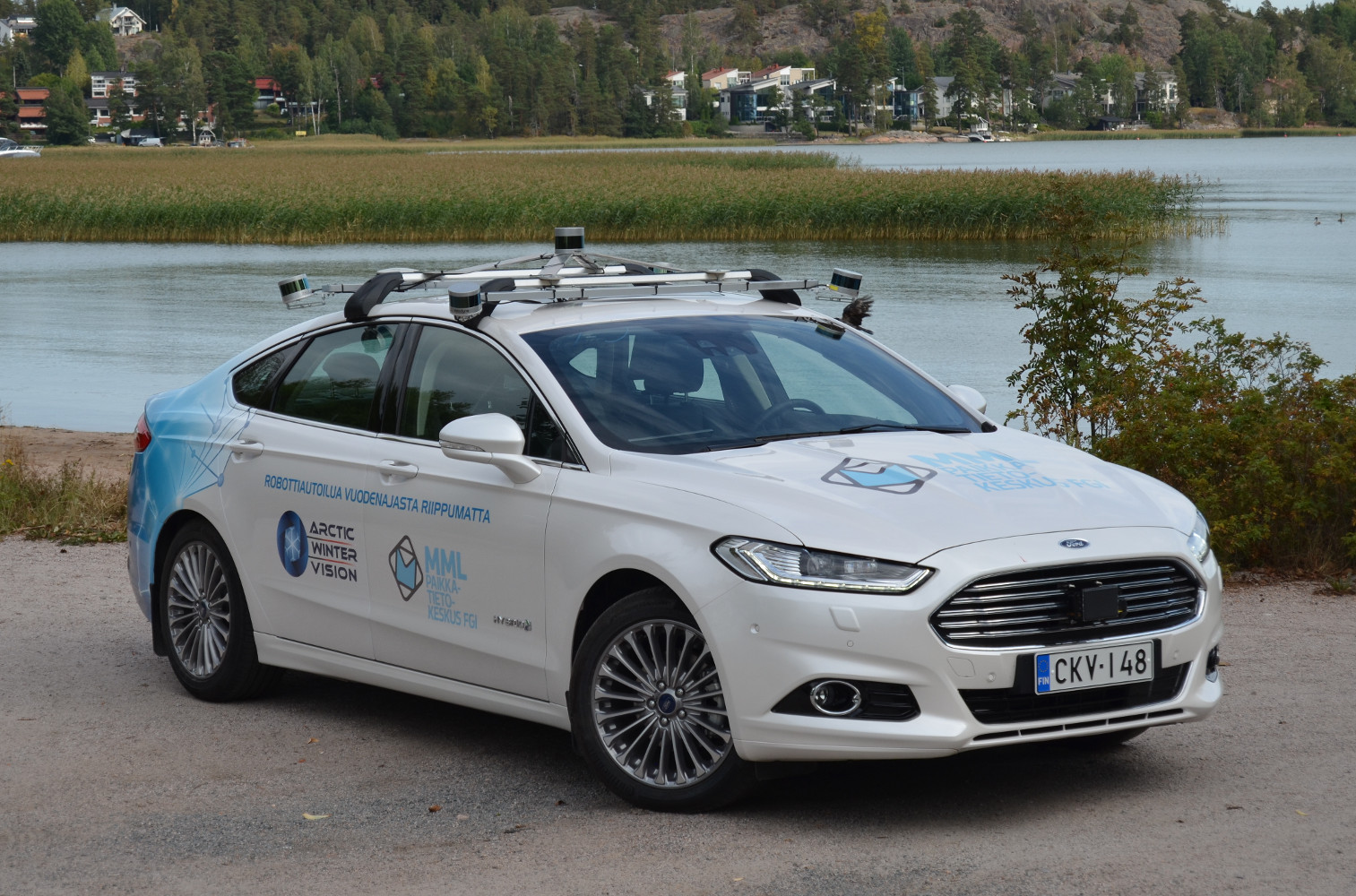}
\caption{FGI autonomous driving research platform ARVO.}
\label{fig:arvo}
\end{figure}

In this article we extend this previous work to a multimodal setting in which the CNN utilizes also lidar data in a sensor fusion manner. We also present further experiments of this method with our research platform ARVO~(Figure~\ref{fig:arvo}) to operate on real-world gravel road tracks and adverse weather conditions. In addition, we compare the performances of models trained on different sensor modalities to gain understanding on the performance reliability based on different sensors. 

The premilinaly results of this work are presented in the master's thesis~\cite{jmaanpaa_dippa}.

\section{Related Work}

The earliest work on end-to-end autonomous steering is the work related to The Autonomous Land Vehicle in a Neural Network (ALVINN)~\cite{NIPS1988_95}, in which a fully connected network with a single hidden layer was trained with low-resolution camera images and laser range measurements. M. Bojarski et al.~\cite{bojarski2016end} extended this early work with modern computational resources, 
utilizing camera images with higher resolution, CNN for steering prediction and a large training dataset. They also presented a VisualBackProp method in~\cite{bojarski2017explaining} for visualizing the operation of the CNN. 

The literature review in~\cite{FengHaase2019multimodal} presents a large listing of current multimodal sensor fusion methods for object detection and semantic segmentation in road environment. Our problem of steering wheel angle prediction differs from these tasks, but the principles of fusing sensor data can be applied in both problem scenarios. Lidar data is fused with RGB camera data in several works related to our problem setting, which include applications in autonomous driving simulations~\cite{xiao2019multimodal, liu2017learning} and indoor navigation~\cite{patel2017sensor, patel2017reducing}. However, as far as we know, none of these studies apply sensor fusion in actual real-world on-road steering task.

Some related work apply end-to-end learning in off-road and gravel road driving conditions with miniature cars~\cite{muller2006off, pan2018agile}. There is also some research focusing on autonomous driving tasks in adverse weather conditions, such as multimodal object detection in poor visibility and sensor failure situations~\cite{bijelic2019seeing, pfeuffer2018optimal} and simulating adverse weather conditions with CycleGAN-based methods for autonomous steering model performance testing~\cite{zhang2018deeproad, machiraju2020little}. Several works also apply novel sensor fusion methods for addressing uncertainty in the multimodal input data~\cite{bijelic2019seeing, valada2018self, liu2017learning}. Otherwise there is less related work on applying deep learning methods for autonomous steering that specifically address adverse road and weather conditions.

There are also several other ways to develop the models utilized in the car control prediction. One could use a time-dependent model with LSTM layers for dynamic model operation~\cite{chi2017deep, du2019jointly} or predict car speed control in addition to the steering wheel angle~\cite{yang2018end}. However, we did not adapt these methods to our model as our problem setting is more related to lane detection in adverse conditions than dynamic route planning and control.

\section{Method}

\begin{figure*}[!t]
\centering
\includegraphics[width=\linewidth]{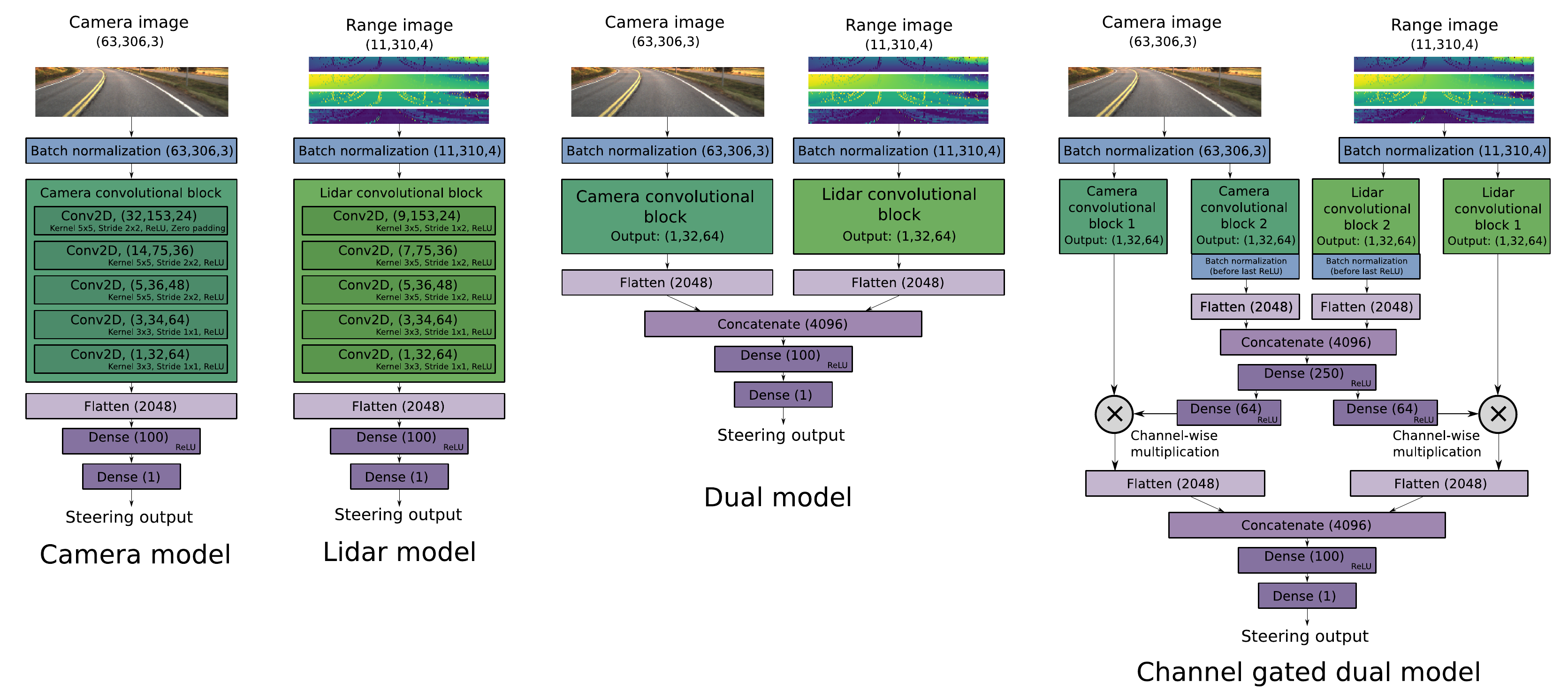}
\caption{Our model architectures based on different sensor modalities. The layer output tensor sizes are shown in each layer with other layer-specific details. The convolutional blocks share the architectures for each modality and their details are shown in the single-modality models. The channel gated dual model has two branches from the normalized inputs: one for gain evaluation and one for steering prediction from gated convolutional block output channels.}
\label{fig:models}
\end{figure*}

Our CNN model architecture designs for the steering wheel angle prediction are based on the PilotNet presented in the work of M. Bojarski et al.~\cite{bojarski2016end}. Our models utilize similarly five convolutional layers for a single modality but incorporate only two dense layers after flattened convolutional part output for steering prediction as utilizing less dense layers resulted in smaller validation error in our preliminary experiments. The operation of the convolutional layers can be seen as feature recognition from the input and the dense part operation is responsible for the control prediction based on the detected features, even though it is possible that both parts of the network contribute to each of these tasks. Both the convolutional layers and dense layers have rectified linear unit (ReLU) activations with the exception of the last layer which has linear activation for output prediction. The convolutional part architectures for a single modality and the dense layer part are similar between all models to improve the consistency in the model performance comparison. We utilized mean squared error of the predicted steering wheel angle as the loss function. 

We trained four CNNs that are based on different sensor modalities. Two first models utilize only one modality and they are called the camera model and the lidar model in this work, corresponding to the utilized sensor. We also trained two sensor fusion architectures that utilize both camera and lidar modalities, called the dual model and the channel gated dual model.
The camera model architecture can be seen as a close equivalent of the PilotNet, whereas the lidar model and the dual model architectures are guided by the design of the PilotNet due to their amount of convolutional layers and channels. The network architectures are shown in Figure~\ref{fig:models}. 

We used two sensor fusion principles with the models utilizing both sensor modalities.
In the first dual model, we used the "middle fusion" principle with concatenation as the fusion operation after convolutional part of the network, as described in the terms of the literature review in~\cite{FengHaase2019multimodal}. This is the simplest fusion operation in this work to gain a baseline performance when no complex fusion architectures are used. In the channel gated dual model there are two parallel dual model architectures, as a gating subnetwork predicts $2\times 64$~gate weights from the normalized input to gate each channel of the two convolutional block outputs in a parallel steering subnetwork which predicts the actual steering wheel angle. The operation of this model can be seen as two parallel tasks: the gating subnetwork evaluates if the features found in the data are reliable and the steering subnetwork operates on these filtered features. Of course, the gating subnetwork might learn to affect the steering output directly and the steering subnetwork might perform some filtering on the input data. 

The output of the utilized model is smoothed during online operation with a moving exponential average with decay parameter $0.9$ prioritizing current model output in the steering actuation. In addition, the output is limited with safeguard criteria which prevent significantly rapid steering maneuvers that do not happen in normal car operation.

\section{Data Collection}

In this section we present our autonomous driving platform with the scope related to this work, our dataset and data preprocessing methods.

\subsection{FGI Autonomous Driving Research Platform ARVO}

Autonomous Driving Research Team in Finnish Geospatial Research Institute FGI has developed an autonomous driving research platform ARVO (Figure~\ref{fig:arvo}) 
for data collection and on-road testing of self-driving methods. Our car is based on Ford Hybrid Mondeo model 2017 and it is equipped with DataSpeed ADAS Kit 
drive-by-wire interface for car control recording and for operating the car with a computer. The current sensor setup of the car includes three front-facing RGB cameras and two front-facing monochromatic cameras, five Velodyne lidars, three radars located at the car front and at the two back corners, 12 ultrasonic sensors and Novatel SPAN-IGM-S1 for GPS and inertial measurements. There are also several computers and a network-attached storage for data management and car operation.

This work utilizes the data from three RGB cameras (PointGrey Chameleon3, CM3-U3-50S5C-CS) and Velodyne VLP-32C lidar in addition to the car control measurements. The lidar is located at the center of the car roof, leveled horizontally. 
The data collection and model operation is performed on Intel NUC (NUC7i7BNH) mini PC with Intel Core i7 processor.

\subsection{Data Collection}

We collected approximately 28 hours of driving data, utilizing 27 hours and 49 minutes for model training and validation and 30 minutes for performance evaluation. The data was collected between September 2018 and May 2019 from Southern Finland and Western Lapland area and it includes 1,020,000~frames obtained with 10~fps frame rate. The distance travelled during the data collection was approximately 1590 kilometers, and the dataset included various different road types and different weather conditions. 52\% of the data was collected in winter conditions, 25\% were classified as gravel roads and 19\% were classified as other roads with no lane markings (these groups of data are not mutually exclusive). Some challenging weather conditions such as snowfall and rain did significantly decrease the data quality in a fraction of the data. Intersections, lane changes and stops were excluded from the dataset as they did not fit the lane-keeping task of the model.

The dataset includes YUV color images with 63$\times$306 resolution, cropping the car hood and the area above horizon level from the image. 
We captured images with three front-facing cameras as side camera images were used for augmenting the car location as in the work of M. Bojarski et al.~\cite{bojarski2016end}, further discussed in the following section. 
The side cameras were located 39~cm from the center camera on each side and also 8~cm higher than the center camera. Therefore the side camera images were corrected with a hand-tuned 3D perspective transform in order to augment their position to the center camera level.

The lidar data is processed as 11$\times$310 range images corresponding spherical coordinates in which each row corresponds a single ring in the lidar scan. Each range image is subsampled from a single lidar scan, covering the vertical view from~$-11.3\degree$ to~$-2.7\degree$ with respect to horizontal level and $\pm34.4\degree$~horizontal view in the forward direction of the car. In the subsampling process each range image pixel contains approximately a single echo from the lidar scan, but in the case of two overlapping echoes the closer echo is selected as the pixel value. We decided to use four channels in the range images, which are the XYZ-coordinates of the sampled echo and the normalized reflectance of the echo. The reflectance values measured by the Velodyne sensor are normalized via averaging the reflectance values of the road echoes in the collected dataset. Range image pixels with no echo have value zero in all four channels. 

As there were less samples from road curves in the collected dataset, the dataset was balanced by including the samples with large steering wheel angle several times in the dataset.
In addition, we applied random color transformations in the camera images, shifting the hue, gamma and saturation values with small random values. This generalized the model to operate in different lighting conditions. We also applied random rescalings on the normalized lidar reflectance values within each row in the lidar range image, as the lidar channels corresponding range image rows still had systematic deviations in reflectance values after the reflectance normalization.
The corresponding steering wheel angle for a single sensor data sample was determined from the frame 0.2~seconds after the current frame, as this corrects the delay in the online model inference and steering actuation.

As the lidar scanning speed has small variance during data collection and online tests, the lidar scans were not accurately synchronized with the camera shutter and the models utilize the closest matching samples captured from camera and lidar for each frame during offline model training and performance evaluation. During online tests the models utilize the most recent samples from each modality for steering evaluation.

There were no specific camera calibrations or lidar movement corrections applied. This might decrease the performance of our models, but we assumed that models could also fit into uncorrected data. This also means that our trained models are dependent on the current physical sensor setup.

\subsection{Data Augmentation}\label{sec:augmentation}

In order to train the steering model to recover back to lane center from displaced positions, we augmented the training data to include samples which simulate the displaced sensor positions if the car was not at the center of the lane as otherwise assumed in the collected data.
The corresponding steering wheel angle is corrected with a hyperparameter 
to train the model to steer the car back to lane center in these augmented samples. The corrected steering angle is evaluated as follows:
\begin{equation}
    \theta = \theta_{\text{orig}} - wd
\end{equation}
Here $\theta$ is the corrected steering wheel angle, $\theta_{\text{orig}}$ is the measured steering wheel angle, $d$~is the displacement from the lane center in meters, and $w$~is the hyperparameter to adjust the amount of steering correction back to the lane center. We used the value of $w=0.52$ in our experiments, based on the observations from the preliminary on-road tests.
We augmented camera data by capturing images from three cameras, similarly as in the work of M. Bojarski et al.~\cite{bojarski2016end}. The lidar position is also augmented to similar displaced positions as the side cameras.

As the lidar scans were always collected at a single location with respect to the car, the lidar range images were augmented with a 3D transformation procedure. The original scan of 3D points is first moved to the augmented position. Then the augmented range image values are interpolated from the 3D points, maintaining lidar sensor firing geometry. The range values are otherwise interpolated linearly unless the radial distance between them is over 3.0~meters, in which case the points are handled belonging to separate objects. The zero echo positions and reflectance values are also maintained in the interpolation process. A more detailed explanation of the lidar augmentation algorithm is available in~\cite{jmaanpaa_dippa}.

\begin{figure}[!t]
\centering
\includegraphics[width=\linewidth]{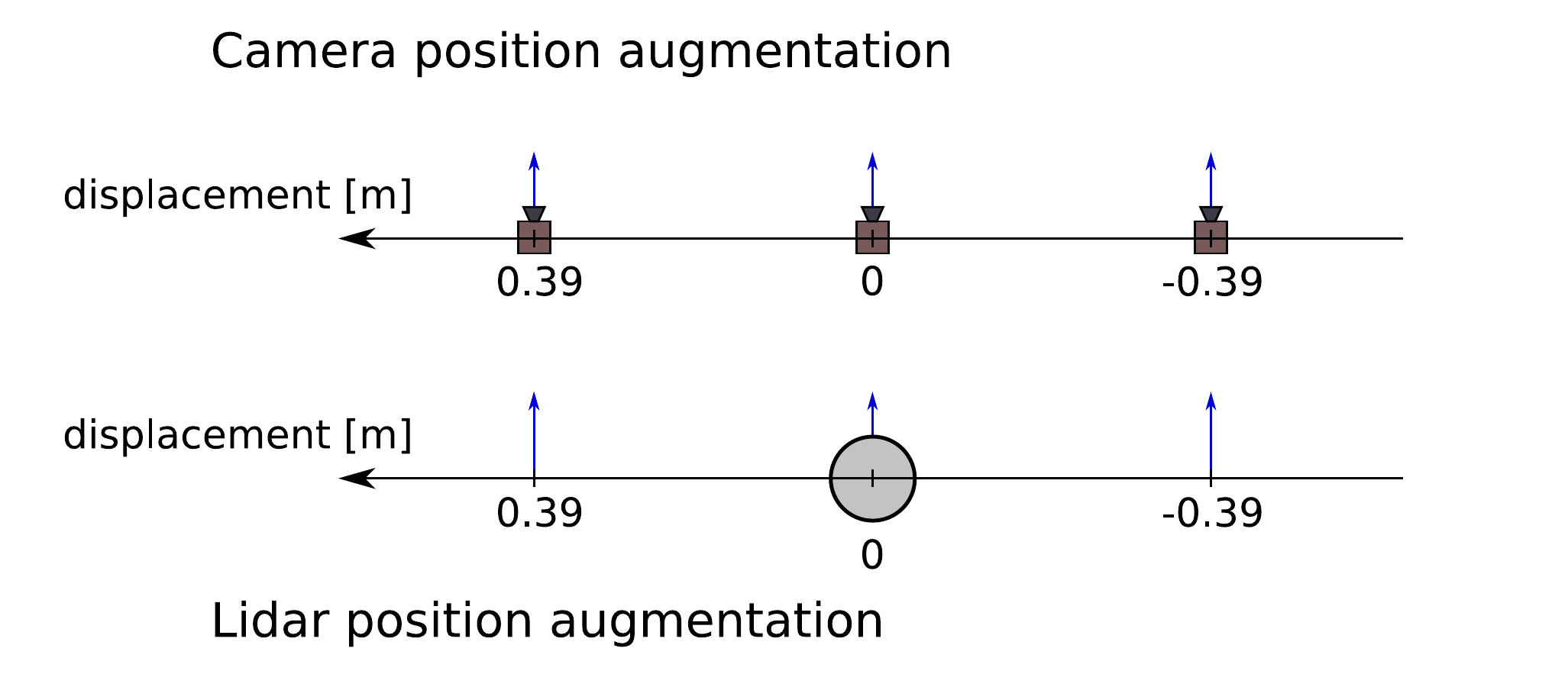}
	\caption{A visualization of the sensor position augmentation. Blue arrows indicate the sensor positions and orientations in the augmented data.}
\label{fig:augmentation}
\end{figure}

The sensor positions are augmented to three positions which correspond to the camera positions. The augmentation setup is illustrated in Figure~\ref{fig:augmentation}.

\section{Experiments}

We trained our models for 12 epochs with Adam optimization algorithm~\cite{kingma2014adam} using learning rate~$10^{-4}$ and otherwise default parameters. We selected the model with lowest error on augmented validation set from each training. The validation error is estimated also on augmented samples as this gives better measure of how the model has learned to response to the actual training data. We used 10\% of the data in the training dataset for model validation 
by selecting 2 second samples between 20 second intervals within the whole training dataset. This reduces the risk for possible overfitting effect in validation set due to the similarity of consecutive samples. 
We trained each model five times and evaluated the validation and test errors for each of them to obtain information on the error distributions for each model. For the test set error evaluation and on-road testing, we selected the model with lowest non-augmented validation error from the five selected models from five trainings as the test set error is evaluated only on non-augmented samples.

\subsection{Validation and Test Errors}

In order to estimate model error numerically on real-world conditions, we measured the prediction error of the models on a test dataset consisting of four individual driving sequences in different road and weather conditions. We utilized the samples from the center camera and the original lidar position for this evaluation as they correspond to a real steering wheel angle, even though this does not measure model steering from displaced positions on the lane. The four sequences contain 15 minutes of a wet asphalt road with lane markings, 4 minutes of a gravel road during fall, 6 minutes of an asphalt road covered with snow and slush and 5 minutes of a gravel road covered entirely with snow. We also report the errors on our validation set for reference, calculated in similar manner.

\begin{figure}[!t]
\captionsetup[subfloat]{farskip=3pt,captionskip=0pt}
\centering
	\subfloat{\includegraphics[width=2.5in]{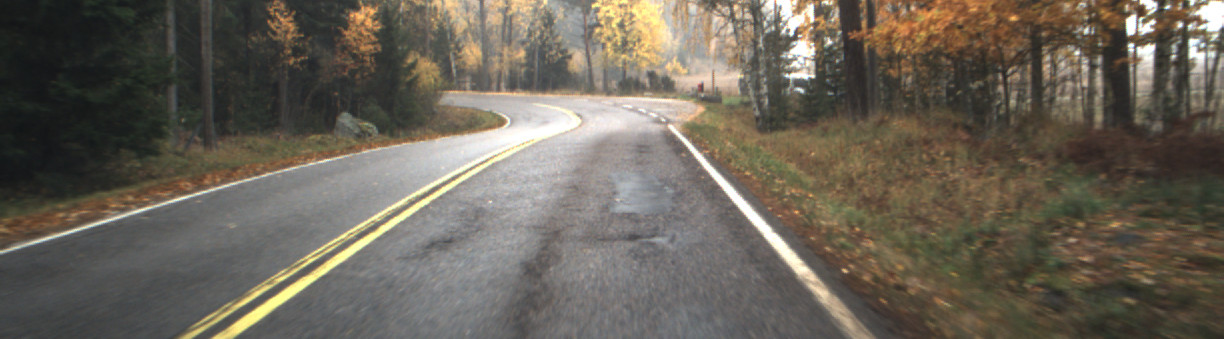}}	
\hfil
	\subfloat{\includegraphics[width=2.5in]{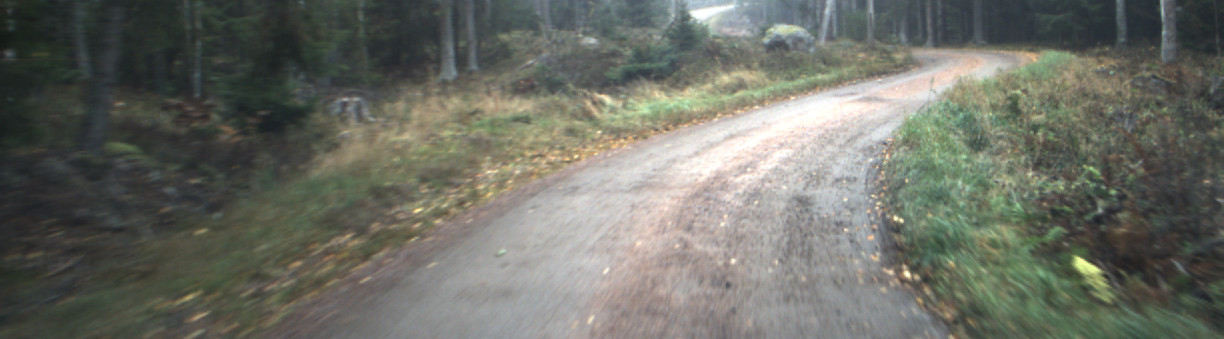}}
\hfil
	\subfloat{\includegraphics[width=2.5in]{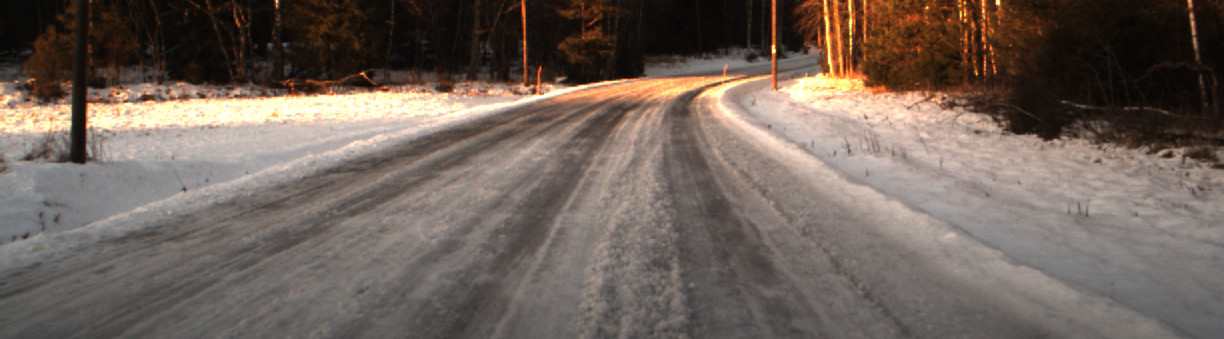}}
\hfil
    \subfloat{\includegraphics[width=2.5in]{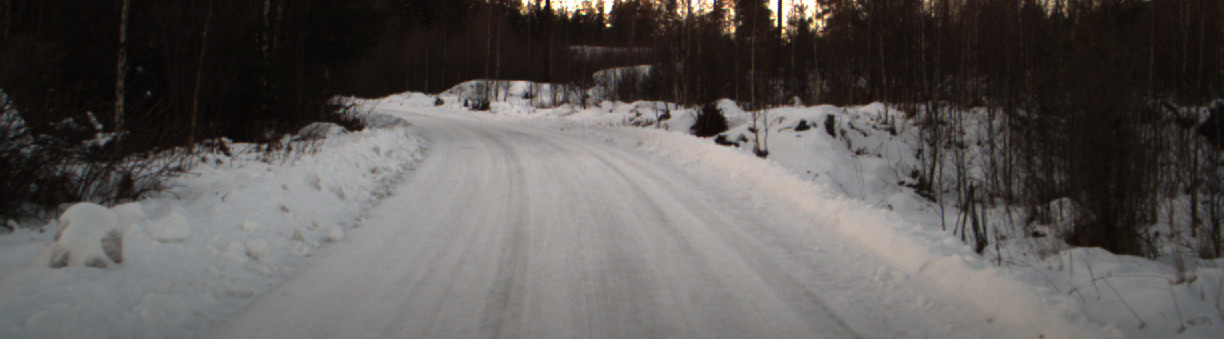}}
\caption{Typical test dataset samples from the four driving sequences.}
\label{fig:testsamples}
\end{figure}

Samples from the four test set driving sequences are illustrated in Figure~\ref{fig:testsamples} and model performances on them are presented in     Table~\ref{tab:rmse}. Specifically, we provide root mean square (RMS) error between the predicted and actual steering wheel angle within the recorded dataset. Furthermore, we also compute the standard deviation of the error between five separately trained similar models. The evaluation performance of the proposed architectures on the validation set is presented in Table~\ref{tab:rmse_validation}.

We notice that the channel gated dual model has the best accuracy in the test set. Its accuracy is clearly better than the accuracy of the dual model, which is due to better sensor fusion architecture. However, this difference is not that clear in the validation set results, which might be due to larger variance of different road types in the validation set. The camera and lidar models have similar accuracy when considering both validation and test set errors, which suggests that they are equally reliable in steering wheel angle prediction. The model accuracy is also clearly improved if both modalities are used as seen from the smaller error of dual models. The systematic difference between validation and test set results are due to different level of difficulty and different magnitude of steering between datasets.

\begin{table}[!t]
\renewcommand{\arraystretch}{1.3}
\caption{Error on validation set.}
\label{tab:rmse_validation}
\centering
	\begin{tabular}{|l|r|r|r|}
		\hline
					 & \multicolumn{1}{l|}{Steering RMSE} & \multicolumn{1}{l|}{Average} & \multicolumn{1}{l|}{Std}   \\ \hline
					 Camera model & $9.41\degree$              & $9.62\degree$ & $0.119\degree$ \\ \hline
					 Lidar model  & $9.24\degree$              & $9.41\degree$ & $0.135\degree$ \\ \hline
					 Dual model   & $8.44\degree$              & $8.52\degree$ & $0.071\degree$ \\ \hline
					 Channel gated dual model & $8.23\degree$  & $8.36\degree$ & $0.111\degree$ \\ \hline
	\end{tabular}
\end{table}

\begin{table}[!t]
\renewcommand{\arraystretch}{1.3}
\caption{Error on recorded test dataset sequences.}
\label{tab:rmse}
\centering
	\begin{tabular}{|l|r|r|r|}
		\hline
					 & \multicolumn{1}{l|}{Steering RMSE} & \multicolumn{1}{l|}{Average} & \multicolumn{1}{l|}{Std}   \\ \hline
					 Camera model & $7.18\degree$              & $7.28\degree$ & $0.083\degree$ \\ \hline
					 Lidar model  & $7.33\degree$              & $7.31\degree$ & $0.069\degree$ \\ \hline
					 Dual model   & $6.75\degree$              & $6.68\degree$ & $0.079\degree$ \\ \hline
				     Channel gated dual model & $6.00\degree$  & $6.07\degree$ & $0.101\degree$ \\ \hline
	\end{tabular}
\end{table}

\subsection{On-Road Tests}

We carried out several on-road tests on public roads to demonstrate the online performances of our steering models.
Most of the tests during model development were qualitative which means that the purpose of the test was to gain user experience on the model behaviour for further development of the models. These tests partly affected the decisions made in the sensor data preprocessing and model training. We also performed quantitative tests with our models to compare the performances between them in challenging road conditions. In these tests, the test roads were driven by utilizing each steering model with the assumption that the road and weather conditions do not significantly change between different test drives. As the amount of quantitative tests is small and their results are strongly dependent from the current environmental conditions, we present these results as an example of performance comparison on certain road conditions.

\begin{figure}[!t]
\centering
\includegraphics[width=\linewidth]{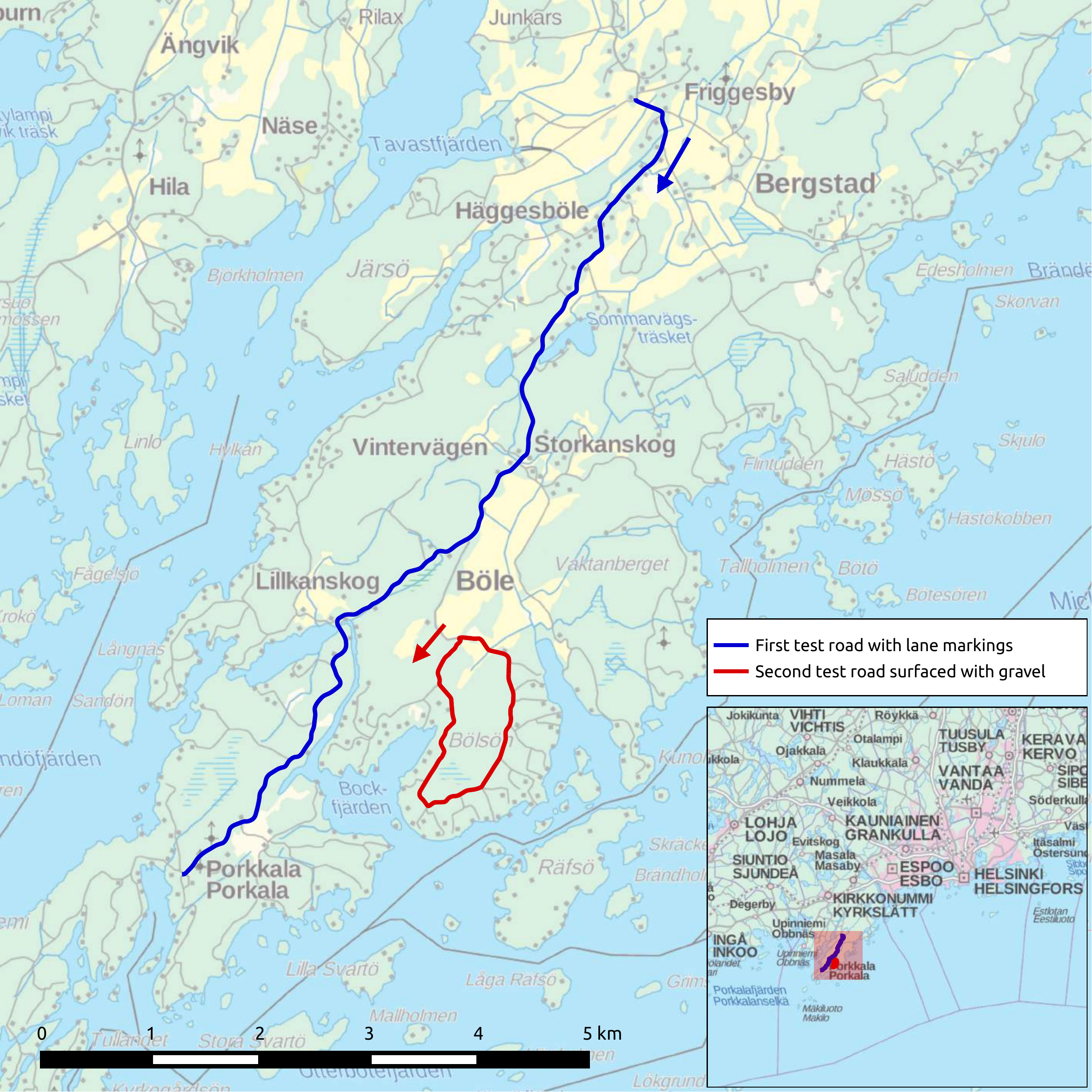}
	\caption{Map of the test roads in Kirkkonummi area, Finland. Blue track is the first test road with partially unclear lane markings and the red track is the second test road surfaced with gravel. Arrows show the direction of the test drives from the start. The background map is from the open data provided by the National Land Survey of Finland.}
\label{fig:testroadmap}
\end{figure}

\begin{figure}[!t]
\captionsetup[subfloat]{farskip=3pt,captionskip=2pt}
\centering
	\subfloat[First test road]{
		\begin{minipage}{2.5in}{
		\includegraphics[width=2.5in]{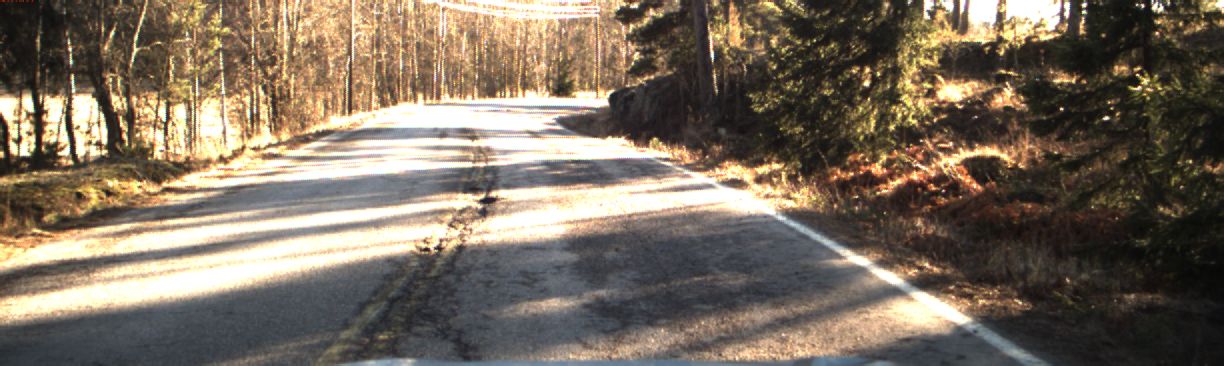}\\[3pt]
		\includegraphics[width=2.5in]{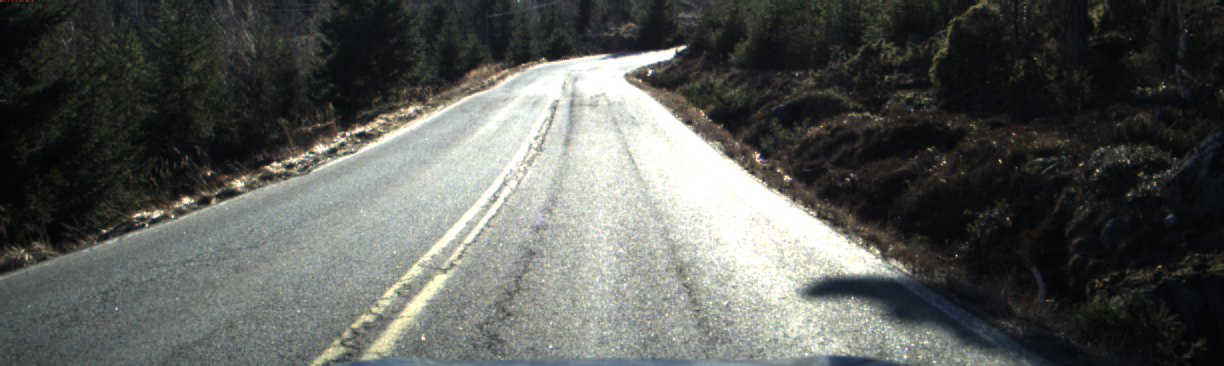}}
		\end{minipage}}
\hfil
	\subfloat[Second test road]{
		\begin{minipage}{2.5in}{
		\includegraphics[width=2.5in]{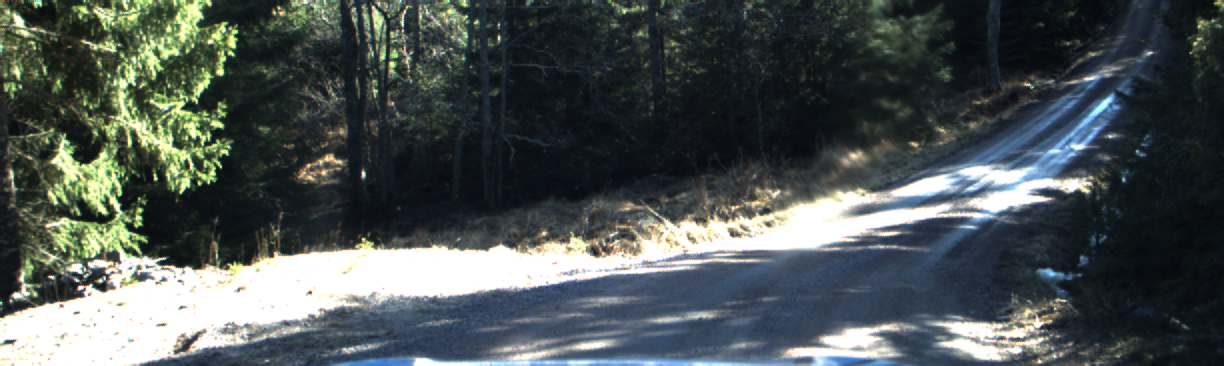}\\[3pt]
		\includegraphics[width=2.5in]{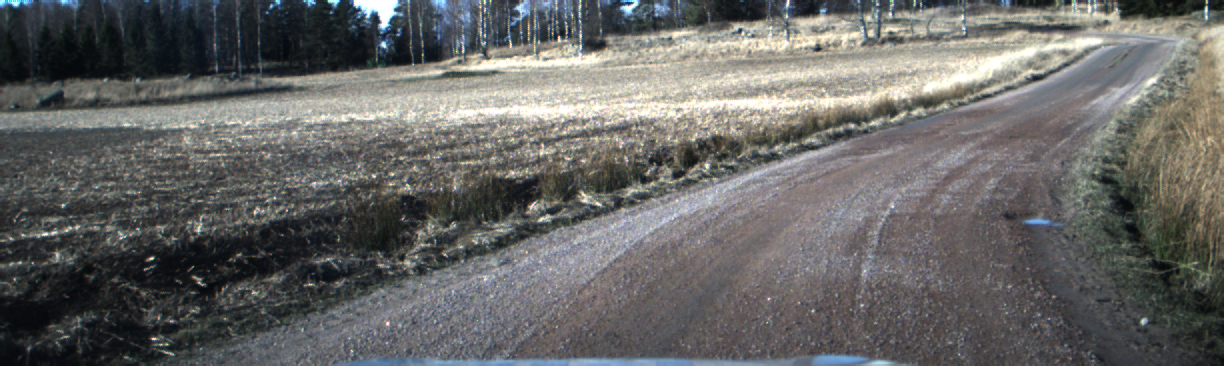}}
		\end{minipage}}
\caption{Typical samples from the test roads collected during the on-road tests.}
\label{fig:onroadtestsamples}
\end{figure}

 We chose two tracks in Kirkkonummi area for the quantitative tests: $9.5$~km track of Porkkalantie which is an asphalt road with partially unclear lane markings and several steep curves, hills, unclear intersections and some frost damage. The second test road is a $4.1$~km single-lane gravel road track with steep hills and curves, consisting of L{\"a}ntinen B{\"o}lentie and It{\"a}inen B{\"o}lentie. A map of the test roads is illustrated in Figure~\ref{fig:testroadmap}. These tests were conducted during March~2020 in clear sunny weather and the car speed was adjusted with a constant velocity controller. The car speed was set to 30~km/h and 40~km/h on the first road depending on the speed limit and 25~km/h speed at the second test road. The speeds were set lower than the speed limits of the roads, as the models operate with a 10~Hz update rate due to the lidar scanning frequency and larger speeds would decrease the model performances due to steering operation latency, which is partly unrelated to model detection accuracy. Typical samples collected from the test roads are visualized in Figure~\ref{fig:onroadtestsamples}.

We recorded two performance metrics during the quantitative tests: the level of autonomy and the amount of driver interventions. The driver was instructed to take over the car control whenever the autonomous steering operation seemed unreliable to keep the car on the current lane.
Each of these corrective actions were counted as a driver intervention. The level of autonomy is defined in a similar way as in the work of~M. Bojarski et al.~\cite{bojarski2016end} which is in the continuous setting
\begin{equation}
	\text{Level of autonomy}~= \frac{\text{Autonomously driven time}}{\text{Operation test time}}~.
\end{equation}
Here the operation test time includes only the test time which is driven with the intention to stay on the current lane, excluding intersections and steering actions related to 
giving way for other traffic.
Due to changing traffic conditions, there were some situations in which other traffic was avoided by driver interventions. As these situations were excluded from the full test time by the above definition, there is an approximately~$1\%$ error in the level of autonomy,
estimated from the amount of test interruption time from the operation test time. In addition, we estimated that there is an approximate error of one driver intervention in the driver interventions metric as the test driver could not tolerate different unreliable steering situations in a systematic manner between different test drives. As sometimes several driver interventions occur within a small time period, we counted manual operation times with less than three seconds of autonomous operation in between as a single driver intervention consisting fully of manual operation time.

\begin{figure}[!t]
\captionsetup[subfloat]{farskip=3pt,captionskip=0pt}
\centering
\subfloat{\includegraphics[width=2.5in]{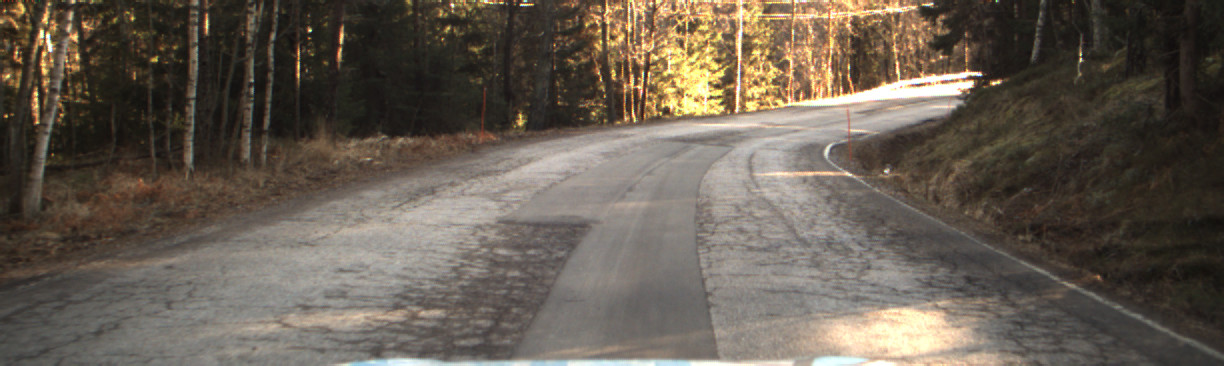}}
\hfil
\subfloat{\includegraphics[width=2.5in]{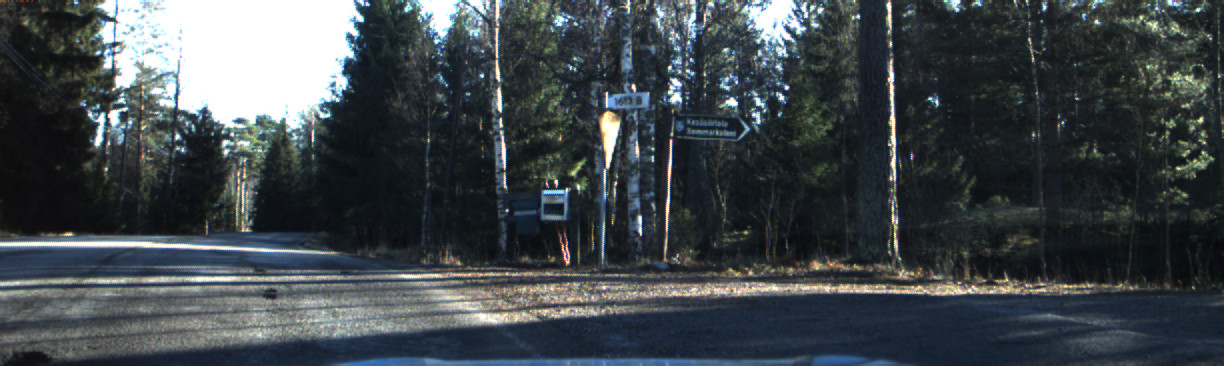}}
\hfil
\subfloat{\includegraphics[width=2.5in]{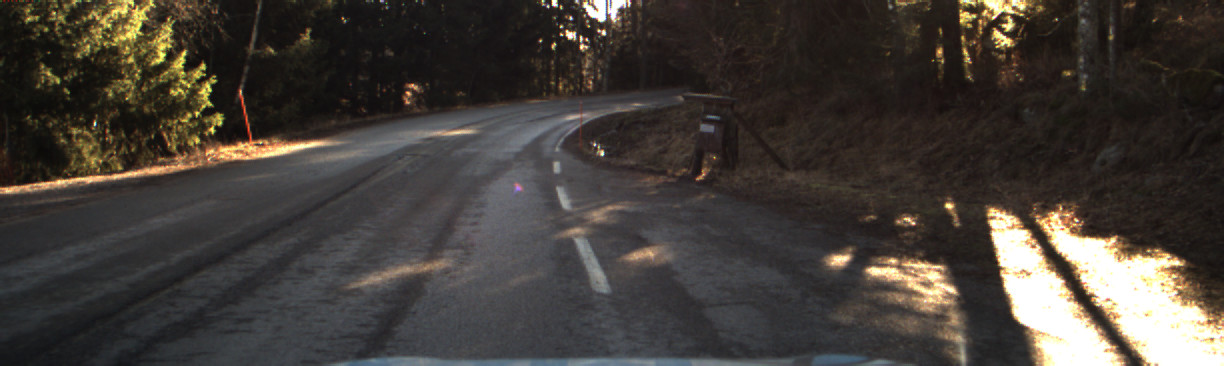}}
\hfil
\subfloat{\includegraphics[width=2.5in]{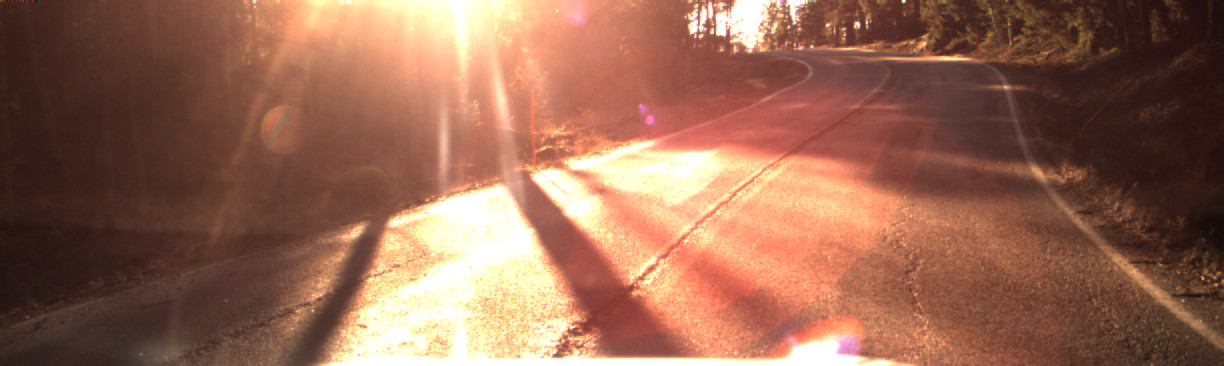}}
\caption{Different driver intervention cases during channel gated dual model operation on the first test road.}
\label{fig:failuresamples}
\end{figure}

The results from the on-road tests on the first test road are provided in Table~\ref{tab:first_onroadtest}. The low sun level during the tests caused significant problems in the operation of the models utilizing camera, as the wet road was easily overexposured in the camera image and there was even direct sunlight to the camera sensor, causing lens flares. These adverse lighting conditions have some variance during tests as the sun position changed during testing, which can affect the comparison of different model results. This can partly explain the contradictory results of the channel gated dual model, which has the best test set error and second lowest level of autonomy. Even though the training data includes samples from overexposure situations, they are not that common that the dual models would have learned to ignore their effect, as seen from the fact that the lidar model has the highest level of autonomy. Better sensor fusion would require more training data from these difficult situations or sensor malfunction augmentation in the training data. However, we observe that utilizing lidar data improves the method reliability as the lidar model and different dual models have better performance than the camera model. The overall insufficient performance of all models is due to the fact that the test road was really difficult with a narrow lane and steep curves. Examples of driver intervention situations on the first test road are shown in Figure~\ref{fig:failuresamples}. 

\begin{table}[!t]
\renewcommand{\arraystretch}{1.3}
	\caption{On-road test results on the first test road (unclear lane markings).}
\label{tab:first_onroadtest}
\centering
	\begin{tabular}{|l|r|r|}
		\hline
					 & \multicolumn{1}{l|}{Level of autonomy} & \multicolumn{1}{l|}{Driver interventions}  \\ \hline
					 Camera model & 92.7\%                            & 21 \\ \hline
					 Lidar model  & 95.1\%                            & 14 \\ \hline
					 Dual model   & 94.4\%                            & 15 \\ \hline
					 Channel gated dual model   & 93.3\%              & 20 \\ \hline
	\end{tabular}
\end{table}

\begin{table}[t]
\renewcommand{\arraystretch}{1.3}
	\caption{On-road test results on the second test road (surfaced with gravel).}
\label{tab:second_onroadtest}
\centering
	\begin{tabular}{|l|r|r|}
		\hline
					 & \multicolumn{1}{l|}{Level of autonomy} & \multicolumn{1}{l|}{Driver interventions}  \\ \hline
					 Camera model & 98.9\%                            & 3 \\ \hline
					 Lidar model  & 100.0\%                           & 0 \\ \hline
					 Dual model   & 100.0\%                           & 0 \\ \hline
					 Channel gated dual model   & 100.0\%             & 0 \\ \hline
\end{tabular}
\end{table}

The results from the on-road tests on the second test road are in Table~\ref{tab:second_onroadtest}. Even though the test road has no lane lines, it was easier for the models as the car speed was lower and the lane was not narrow. Two intersections on the test road were excluded from the test result evaluations as the lane-following task was not straightforward in them even for a human driver. All of the models utilizing lidar data succeeded through the test with full autonomy but the camera model required 3 driver interventions mostly due to problems related to image overexposure.

Documentation from the other tests in several adverse road and weather conditions, including winter conditions, can be seen in a video\footnote{\url{https://www.youtube.com/watch?v=wKrj7cSBKfE}}. Some of the shown on-road tests are performed with earlier versions of our models.

\begin{figure}[t!]
\captionsetup[subfloat]{farskip=3pt,captionskip=2pt}
\centering
	\subfloat[~]{
		\begin{minipage}[t]{2.9in}
		\raisebox{15pt}{1) } \includegraphics[width=2.5in]{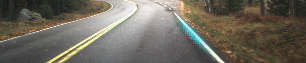}\\[3pt]
		\raisebox{15pt}{2) } \includegraphics[width=2.5in]{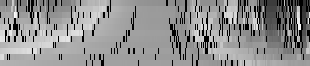}\\[3pt]
		\raisebox{15pt}{3) } \includegraphics[width=2.5in]{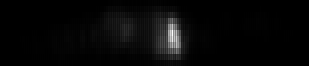}
		\end{minipage}}
\hfil
	\subfloat[~]{
		\begin{minipage}[t]{2.9in}
		\raisebox{15pt}{1) } \includegraphics[width=2.5in]{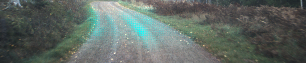}\\[3pt]
		\raisebox{15pt}{2) } \includegraphics[width=2.5in]{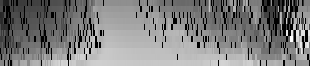}\\[3pt]
		\raisebox{15pt}{3) } \includegraphics[width=2.5in]{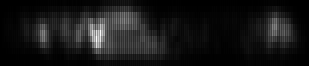}
		\end{minipage}}
\hfil
	\subfloat[~]{
		\begin{minipage}[t]{2.9in}
		\raisebox{15pt}{1) } \includegraphics[width=2.5in]{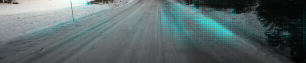}\\[3pt]
		\raisebox{15pt}{2) } \includegraphics[width=2.5in]{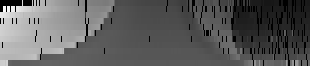}\\[3pt]
		\raisebox{15pt}{3) } \includegraphics[width=2.5in]{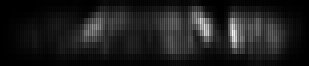}
		\end{minipage}}
\hfil
	\subfloat[~]{
		\begin{minipage}[t]{2.9in}
		\raisebox{15pt}{1) } \includegraphics[width=2.5in]{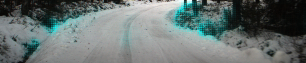}\\[3pt]
		\raisebox{15pt}{2) } \includegraphics[width=2.5in]{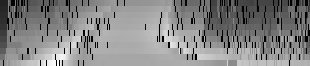}\\[3pt]
		\raisebox{15pt}{3) } \includegraphics[width=2.5in]{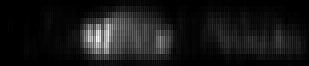}
		\end{minipage}}
\caption{Example VisualBackProp masks evaluated on samples from the four test set sequences (a-d). Camera VisualBackProp mask is shown in cyan overlay color on camera images (1) and the range image $z$-channel (2) is shown to help in the interpretation of the range image VisualBackProp mask (3). The masks are evaluated from the steering subnetwork of the channel gated dual model considering the gated output of the convolutional blocks.} 
\label{fig:vbp}
\end{figure}

\subsection{Channel gated dual model operation visualization with VisualBackProp method}

In addition to the model performance experiments, we implemented the VisualBackProp method as described in~\cite{bojarski2017explaining} to visualize the channel gated dual model operation within different road conditions. The VisualBackProp method visualizes the areas in the input image data that cause activation in the last convolutional layer. This is done by averaging all output channels in each convolutional layer and upsampling the averaged output of the last layer to the previous layer averaged output size and multiplying these elementwise. This upsampling and multiplication is repeated to obtain a single activation mask on the CNN input image. As the channel gated dual model had several CNN branches, the VisualBackProp masks were evaluated for both camera and lidar convolutional blocks in the steering subnetwork, starting from the gated output of the convolutional blocks to include the effect of gating in the mask evaluation. The mask values are scaled logarithmically in our visualizations in order to show the weaker activations as well as the strongest activations.

Example VisualBackProp masks on test dataset samples are shown in Figure~\ref{fig:vbp}. We observe that lane lines (especially the right lane line) and other road features corresponding road curvature cause activation in the camera convolutional block layers. These features have usually high contrast differences, and therefore some features that have this behaviour and do not correlate with the road curvature can also cause activation in the convolutional layers, which is partly seen in the last camera image example in Figure~\ref{fig:vbp}.

The VisualBackProp masks on range image samples in Figure~\ref{fig:vbp} show that the lidar convolutional block can detect the right lane line, road shoulders, the road itself and some objects outside the road. The lane lines can be observed from the lidar data as usually there are high reflectance values and zero echoes in their vicinity. However, these masks are difficult to interpret due to the low resolution of lidar range images.

\section{Conclusion}

In this paper, we propose a CNN-based approach for the problem of the steering angle estimation in adverse road and weather conditions. The proposed architecture using both camera and lidar data demonstrates good performance in challenging road and weather conditions. Utilizing lidar modality in addition to camera modality increases the accuracy and reliability of our approach, especially when the camera modality suffers from bad data quality. These observations are supported by several on-road experiments. However, there are some challenging situations that require better training data and more accurate sensor fusion. Furthermore, it is difficult to scale on-road testing safely without accurate simulations. We leave these potential research directions for future work.

\section*{Acknowledgment}

Academy of Finland projects (decisions 319011 and 318437) are gratefully acknowledged.

Authors' contribution is the following: Maanp{\"a}{\"a} designed and performed the experiments and wrote the manuscript. Maanp{\"a}{\"a}, Taher, Manninen, and Pakola took equal shares in instrumenting the autonomous driving platform and developing software. Maanp{\"a}{\"a}, Taher, and Pakola participated to data collection. Melekhov advised in the model development and provided feedback on the manuscript. Hyypp{\"a} supervised the project.

In addition the authors would like to thank Heikki Hyyti, Antero Kukko and Harri Kaartinen from the Finnish Geospatial Research Institute FGI and Juho Kannala from Aalto University School of Science for assistance and advice during this work.



\bibliographystyle{IEEEtran}
\bibliography{IEEEabrv,./bibliography}
%
%
%

\end{document}